\documentclass[conference]{IEEEtran}
\usepackage{subfig}
\usepackage{array}
\usepackage{longtable}
\usepackage[flushleft]{threeparttable}
\usepackage{morefloats}
\usepackage{mwe}
\usepackage{verbatim}
\usepackage{multirow}
\usepackage{amsmath}
\usepackage{multicol}
\usepackage[english]{babel}
\usepackage{amssymb}
\begin{document}
\title{Classifying Documents within Multiple Hierarchical Datasets using Multi-Task Learning}

\author{\IEEEauthorblockN{Azad Naik}
\IEEEauthorblockA{Department of Computer Science\\
George Mason University\\
Fairfax, VA 22030\\
Email: anaik3@gmu.edu}
\and
\IEEEauthorblockN{Anveshi Charuvaka}
\IEEEauthorblockA{Department of Computer Science\\
George Mason University\\
Fairfax, VA 22030\\
Email: acharuva@gmu.edu}
\and
\IEEEauthorblockN{Huzefa Rangwala}
\IEEEauthorblockA{Department of Computer Science\\
George Mason University\\
Fairfax, VA 22030\\
Email: rangwala@cs.gmu.edu}}

\maketitle

\begin{abstract}

Multi-task learning (MTL) is a 
supervised learning paradigm in which the prediction 
models for  several
related tasks are learned jointly to achieve 
better generalization performance. When there are only a few training examples per task, MTL considerably outperforms
the traditional Single task learning (STL) in terms of prediction accuracy.
In this work we 
develop an MTL based approach for classifying documents that are archived within 
dual concept hierarchies, namely,  DMOZ and Wikipedia.
We solve the multi-class classification problem by defining one-versus-rest
binary classification tasks for each of the different 
classes across the two hierarchical datasets. Instead of learning a linear 
discriminant for each of the different tasks independently, we 
use a MTL approach with relationships between the different 
tasks across the datasets established using the non-parametric, lazy, 
nearest neighbor approach.
We also develop and evaluate a transfer learning (TL) approach and 
compare the MTL (and TL) methods against the standard single task 
learning and semi-supervised learning approaches.
Our empirical results demonstrate the strength of our developed
methods that show an improvement especially when there are 
fewer number of training examples per classification task.

\end{abstract}


\section{Introduction}

Several websites like Wikipedia, DMOZ and Yahoo archive documents (text data) into hierarchies with 
large number of classes. Several classification methods have 
been developed to automatically classify text documents into different classes.
In this work we seek to leverage the often implicit 
relationships that 
exists between multiple archival datasets to classify 
documents within them in a combined manner. Further, datasets have
several classes with very few samples which make it harder to learn 
good classification models.

Specifically,  we develop a  Multi-Task Learning (MTL) based approach 
to learn the model vectors associated with
several linear classifiers (one per class) in a joint fashion. Using the 
nearest-neighbor algorithm we identify the hidden relationships between 
the different document datasets and use that within the MTL framework.
MTL approaches are known to achieve superior performance on 
unseen test examples, especially 
when the number of training examples is small. 
MTL has been successfully
applied in varied applications such as
medical informatics \cite{30}, structural 
classification \cite{1}, sequence analysis 
\cite{12}, web image and video search \cite{48}.

In this paper our 
key contributions include development of a document classification method using 
MTL that leverages information present across dual hierarchical datasets. We focused on 
classifying documents within classes as categorized by
Wikipedia and DMOZ dataset. 
In text classification, for 
each of the class 
labels we define a
binary one-versus-rest 
classification task. We then find the related tasks corresponding
to each task using $k$-nearest neighbor, which 
is then learned together to find 
the best suited model vector (or parameters)  corresponding to each task.
Based on how information
from related tasks 
was integrated with the original classification task we 
developed two class of approaches: (i) Neighborhood Pooling Approach and
(ii) Individual Neighborhood Approach. We evaluated the performance
of our MTL approach for document classification with dual hierarchical datasets 
against a transfer learning approach, semi-supervised learning 
approach and a single task learning approach.
Our empirical evaluation demonstrated merits of the MTL approach in 
terms of the classification performance.

The rest of the paper is
organized as follows. Section II provides background related to MTL and TL. Section III 
discusses our developed methods. Section IV provides the experimental protocols. Section V discusses 
the experimental results. Finally, Section VI draws conclusion
and provides several future directions. 

\section{Background}

\subsection{Multi-Task Learning}

Multi-Task Learning (MTL) \cite{7} is a rapidly growing machine learning 
paradigm that involves 
the simultaneous training of multiple, related prediction tasks. This 
differs from the traditional single-task learning (STL), where for 
each task the model is learned independently. MTL-based 
models have the following advantages: (i) they leverage the training 
signal across related tasks, which leads to better generalization ability
for the different tasks and (ii) empirically they have been shown to outperform 
  STL models, especially when there are few examples per task and the 
  tasks are related \cite{7}\cite{8}\cite{9}\cite{10}. 
The past few years has seen an tremendous growth in the development 
and application of MTL-based approaches. A concise review of all these 
approaches can be found in the survey by Zhou et. al. \cite{46}.

For STL, we are given a training set with $n$ examples. Given
an input domain ($\mathcal{X}$) and output domain ($\mathcal{Y}$), the 
$i$-th training 
example is represented by a pair $(x_i, y_i)$ 
where $x_i \in \mathcal{X}$ and $y_i \in \mathcal{Y}$.
Within classical machine learning,
we seek to learn a mapping function 
$f: \mathcal{X} \in \mathbb{R}^{d} \rightarrow \mathcal{Y}$, where 
$d$ denotes the dimensionality of the input space.
Assuming that $f(x)$ is a 
linear discriminant function, it is 
defined as   $f(x) = sign (\langle \theta, x \rangle + c)$,
where $\theta \in \mathbb{R}^{d}$ denotes the weight 
vector or model parameters. 
$f(x)$   allows us to make predictions 
for new and  unseen examples within the $\mathcal{X}$ domain.
This model parameter $\theta$ are learned by  minimizing a loss 
function across all the training examples,
while restricting the model to have
  low complexity using a regularization penalty. As such, the 
STL objective can be shown as: 

\begin{equation} \min_{\theta} \sum_{i=1}^{n} \underbrace { L(\theta, x_i, y_i)}_{Loss}  + \lambda \underbrace{R(\theta)}_{Regularization} \end{equation}
       
where  $L(\cdot)$ represents the loss function 
being minimized, $R(\cdot)$ represents the 
regularizer (e.g., $l_1$-norm) 
and $\lambda$ is a parameter that balances
the trade-off between the loss function 
and regularization penalty. 
The regularization term safeguards against model over-fitting and 
allows the model to generalize to the examples not 
encountered in the training set.

Within MTL, we are given $T$ tasks with training examples
per task. For the $t$-th task, there are $n_t$ number of training examples that
are represented by $\{(x_{it},y_{it}) ; \forall i=1\ldots n_t \}$.
We seek to learn $T$ linear discriminant functions, each 
represented by weight vector $\theta_{t}$. We denote the 
combination of all task-related weight vectors as 
a matrix that stacks all the weight vectors as columns, 
$\Theta = [\theta_1,\ldots,\theta_T]$ of dimensions $d \times T$, where $d$ is the 
number of input dimensions.
The MTL objective is given by 
\begin{equation}    \min_{\Theta} \sum_{t=1}^{T}  \sum_{i=1}^{n_t}  \underbrace { L(\theta_t, x_{it}, y_{it})}_{loss} + \lambda \underbrace{R(\Theta)}_{Regularization}  \end{equation}
The regularization term $R(\Theta)$ captures the relationships between the different tasks. Different MTL approaches vary in the
way the combined regularization is performed but most methods seek to 
leverage the ``task relationships'' and enforce the constraint that  
the model parameters (weight vectors) for related tasks are similar.

In the work of 
Evgeniou et. al \cite{11} the model for each task 
is constrained to be close to the average of all the 
tasks.
In multi-task feature learning and 
feature selection methods \cite{13}\cite{14}\cite{15}\cite{16}, sparse 
learning based on lasso \cite{17}, is performed to select or 
learn a common set of features across many related tasks. However, 
a common assumption made by these approaches  \cite{11}\cite{18}\cite{19} is 
that all tasks are equally related. This assumption
does not hold in all cases, especially when there is no knowledge
of task relationships.

Kato et. al \cite{20} 
and Evgeniou et. al \cite{21} propose formulations
which use an externally provided task network or graph 
structure. However, these relationships might 
not be available and may need to be determined 
from the data. Clustered multi-task learning
approaches assume that tasks exhibit a group-wise
structure, which is not known a-priori and seeks to learn 
the clustering of tasks that are then
learned together \cite{22}\cite{23}\cite{24}. Another set 
of approaches, mostly based on Gaussian Process models, learn 
the task co-variance structure \cite{25}\cite{26} and are 
able to take advantage of both positive and negative 
correlations between the tasks.

In this paper
we have focused on the use of 
MTL based models for the purpose of multi-class 
text classification, when the documents are categorized by multiple hierarchical datasets. 
We use a non-parametric, lazy approach to find the 
related tasks within  different domain datasets and use these relationships
within the regularized MTL approach.

\subsection{Transfer Learning}
\label{ref:tl:background}

Related to MTL are approaches developed for Transfer Learning (TL). 
Within the TL paradigm, it is assumed that there exists one/more 
target (or parent) tasks along with previously learned models for 
related tasks (referred to as children/source tasks). While learning
the model parameters for the target task, TL approaches seek to 
transfer information from the parameters of the source tasks.
The key 
intuition behind using TL 
is that the information contained in the source
tasks can help in learning predictive models
for the target task of interest. When
  transferred parameters from 
  the source task assist in better learning the predictive 
  models of the target task then it is referred to as 
  positive transfer. However, in some cases if 
  source  task(s) are not related to the target task, then 
  the  
  TL approach leads to worse prediction performance. This 
  type of transfer is known as negative transfer. It 
  has been shown in the work of Pan et. al \cite{47} that 
  TL improves the generalization performance of 
  the predictive models, provided the source tasks are related to the target 
  tasks.
One of the key differences between TL and MTL 
approaches is that within the MTL approaches, all 
the task parameters are learned simultaneously, whereas 
in TL approaches, first the parameters of the source tasks 
are learned and then they are transferred during the learning 
of parameters for the target task. In the literature, 
TL has also been referred 
to as Asymmetric Multi-Task Learning because of the 
focus on one/more of the target tasks.

Given a target task with $n_t$ training examples, represented 
as $\{(x_{1t},y_{1t}),\ldots,(x_{nt},y_{nt})$ we seek to learn the parameters 
for the target task ($\theta_{t}$) using the parameters from 
  the source tasks given by ($\Theta_{s}$). Using a similar notation 
  as used before the matrix $\Theta_{s}$ represents all the parameters
  from the different source tasks that are learned separately beforehand.
We can write the minimization function for the target task within the TL 
framework as: 
\begin{equation} \min_{\theta_t} \sum_{i=1}^{n_t} \underbrace { L(\theta_t, x_{it}, y_{it})}_{loss}  +  \underbrace{\lambda_1 R(\theta_t) + \lambda_2 R(\theta_t,  \Theta_s)}_{Regularization}\end{equation} 

where the regularization term $R(\theta_t)$ controls model complexity 
of the target task $t$ and the term 
$R(\theta_t, \Theta_s)$ captures how the 
parameters from the source tasks
will be transferred to the target task.
The exact implementation
of the regularization term is discussed 
in 
Section \ref{sec:methodology}.

\section{Methods}
\label{sec:methodology}

Given two different datasets that categorize/archive text documents
(e.g., Wikipedia and DMOZ), our
primary objective is to classify new documents into classes within 
these datasets. We specifically, use regularized MTL approaches to improve
the document classification performance. First, we assume that each of 
the classes within the different datasets is associated with a binary 
classification task. For each of the tasks within one of the datasets we 
want to determine related tasks within the other database, and by performing
the joint learning using  MTL,  we gain improvement in the classification
performance. We compare the MTL approach against the standard STL approach, the 
TL approach that assumes the tasks in one of the datasets to be the
target task and tasks from the second dataset as the source tasks. We also 
compare our approach to a semi-supervised learning approach (SSL). 

\subsection{Finding Related Tasks}

We first discuss our approach 
to 
determine task relationships across the two datasets using 
the non-parametric, lazy nearest neighbor approach (kNN) \cite{42}.
We use kNN 
to find similar classes between the two datasets i.e., Wikipedia and DMOZ.
For determining the 
nearest neighbor(s), we represent each of the classes
within the DMOZ and Wikipedia datasets by their centroidal
vectors. The centroidal 
vector per class is computed by taking the average across all the examples within 
a given class. 
We then use Tanimoto Similarity (Jaccard index)
to compute similarities between the different classes across the two datasets.
Tanimoto 
  similarity is the ratio of the size of intersection divided by the union. The 
  similarity is known to work well for large dimensions with a lot of zeros (sparsity).
Using kNN, we find for each class of interest a set of 
neighboring classes within 
the second dataset. Within the MTL approach we constrain the related
classes to learn similar weight vectors when jointly learning the model
parameters. In TL approach we learn the weight vectors for related 
classes and transfer the information over to the target task. We also 
use the related classes to supplement the number of positive examples
for  each of the classes within a baseline 
  semi-supervised learning approach (SSL).

\subsection{MTL method}
Given the two dataset sources $S_1$ and $S_2$, we 
represent the total number of classes (and hence the number of 
binary classification tasks) within each of them
 by $T^{S_1}$ and $T^{S_2}$, respectively.
The individual parameters per classification task is represented by
$\theta$ with model parameters for $S_1$ denoted by $\theta^{S_1}$ and 
parameters for $S_2$ denoted by $\theta^{S_2}$. The combined model parameters 
for $S_1$ and $S_2$ are 
denoted by $\Theta^{S_1}$ and $\Theta^{S_2}$, respectively.
The MTL minimization objective can then be given by:
\begin{multline} \sum_{t=1}^{T^{S_1}} \sum_{i=1}^{n_t} L(\theta_t^{S_1}, x_{it}, y_{it}) + \sum_{t=1}^{T^{S_2}} \sum_{i=1}^{n_t} L(\theta_t^{S_2}, x_{it}, y_{it}) +\\ 
\lambda_1 \sum_{t=1}^{T^{S_1}} ||\theta_t^{S_1}||_2^2 + \lambda_2 \sum_{t=1}^{T^{S_2}} ||\theta_t^{S_2}||_2^2 + 
\lambda_3 R(\Theta^{S_1}, \Theta^{S_2})  \end{multline}
where the first two terms are loss computed for each of 
the two dataset-specific models. To control the model complexity we then 
include a $l_2$-norm (denoted by $||\cdot ||_2$), 
for each of the different classification tasks within $S_1$ and $S_2$. Finally, 
$R(\Theta^{S_1}, \Theta^{S_2})$ controls the relationships between the tasks 
found to be related using the kNN approach across the two databases. Parameters 
$\lambda_1$, $\lambda_2$ and $\lambda_3$ control the weights associated with 
each of the different regularization parameters. 
Based on how we constrain the related tasks, we 
discuss two approaches: 
\begin{itemize}
\item  Neighborhood Pooling Approach (NPA). In 
this approach,  for each of the
tasks within $S_1$  we find the $k$-related 
neighbors from the other dataset $S_2$. We repeat this by finding 
related neighbors in $S_1$ for each class in $S_2$. 
Then we pool all the training examples within the related classes 
and assume that there exists one pooled task for each of the original task. We then  
constrain the model vectors 
for each task to be similar to the pooled model vector. We represent this 
 as
\begin{small}
\begin{equation}
R(\Theta^{S_1}, \Theta^{S_2}) = \sum_{t=1}^{T^{S_1}} ||\theta_t^{S_1}  - \theta_{NPA(t)}^{S_2}||_2^2 +  \sum_{t=1}^{T^{S_2}} ||\theta_t^{S_2}  - \theta_{NPA(t)}^{S_1}||_2^2 \end{equation}
\end{small}
where $\theta_{NPA}(t)^{S_2}$ represents 
the pooled related neighbor model within $S_2$. The 
weight vectors for the original tasks and new pooled tasks
are learned simultaneously. We
denote this approach as MTL-NPA.

\item Individual Neighborhood Approach (INA). In this 
approach we consider  all the 
$k$ related 
neighbors from the second source 
as individual tasks. As such we constrain each task model vector to be similar
to each of the $k$ related task vectors. The regularization term 
can then be given by:
\begin{small}
\begin{equation}
R(\Theta^{S_1}, \Theta^{S_2}) = \sum_{t=1}^{T^{S_1}}  \sum_{l=1}^{k}||\theta_t^{S_1} -\Theta_{I(l)}^{S_2}||_2^2  + 
\sum_{t=1}^{T^{S_2}}  \sum_{l=1}^{k} ||\theta_t^{S_2} - \Theta_{I(l)}^{S_1}||_2^2 
\end{equation}
\end{small}
where $I(\cdot)$ is an indicator function representing the identified 
nearest neighbor task vectors within the second dataset. 
We refer to this approach as MTL-INA.
\end{itemize}

\subsection{Transfer Learning Approach.} 

The TL 
method differs from the MTL method in the 
learning process. In MTL 
all the related task model parameters are
learned simultaneously whereas
in TL method learned parameters of the related task 
are transferred to the main task to improve
model performance. Within our 
TL method, we 
use the parameters from the 
neighboring tasks within the 
regularization term for the 
main task. 
Similar to the MTL models, we implement both the pooling and individual 
neighborhood approaches for transfer learning.

\subsubsection{Neighborhood Pooling Approach (TL-NPA)}
This method pools the 
$k$ neighbors for each of the tasks considered to be within the primary
dataset. 
After pooling, at first using STL the parameters for the pooled model are learned. The pooled 
parameters for task $t$ from the secondary source database (S) 
are represented as $\theta_{NPA(t)}^{S}$.
Assuming $S_1$ to be the main task dataset we can write the objective function
for each of the $t$ task within $S_1$ as follows:
  \begin{equation}
\min_{\theta_t^{S_1}} \sum_{i=1}^{n_t} L(\theta_t^{S_1}, x_{it}, y_{it}) + \lambda_1 ||\theta_t^{S_1}||_2^2 + \lambda_2 || \theta_t - \theta_{NPA(t)}^{S_2}||_2^2 
  \end{equation}
We can similarly write the objective assuming $S_2$ to be the main/primary dataset.

\subsubsection{Individual Neighborhood Approach (TL-INA)}
In this approach, for 
each of the $k$ neighborhood tasks, we learn the parameter vectors individually (using STL). After this, a transfer 
of information is performed from each of the related tasks to the main/parent tasks. We can represent this within the 
TL objective as follows:
\begin{equation}
\min_{\theta_t^{S_1}} \sum_{i=1}^{n_t} L(\theta_t^{S_1}, x_{it}, y_{it}) + \lambda_1 ||\theta_t^{S_1}||_2^2 + \lambda_2 \sum_{l=1}^{k} || \theta_t - \Theta_{I(l)}^{S_2}||_2^2 
\end{equation}
The last regularization term is similar to the MTL-INA approach discussed earlier, where $I(\cdot)$ represents an indicator 
function to extract the $k$-related tasks. We can also assume $S_2$ to be the target/parent dataset.

\subsection{Single Task Learning}

Single Task Learning (STL) lies within the standard machine learning 
paradigm, where 
each classification task is treated 
independently of each other during the training phase. 
The learning objective of the 
regularized STL model is given by Equation 1. 
In this paper, 
logistic regression is used as the loss function for all the binary
classification tasks. 
One advantage of 
using this loss function is that it is smooth and convex. 

Specifically, the STL objective 
can
be rewritten as,
\begin{equation} \min \sb \theta \sum_{i=1}^{n}log\Big({1+exp(-y_i\theta^{T}x_i)}\Big) + \frac{\lambda}{2}{||\theta||}_2^2\end{equation}
where, $y \in {\{\pm1\}}$ is the binary class label 
for $x$, $\theta$ is the model vector/parameters.
For preventing the model from over-fitting we have used
the $l_2$-norm over the $\theta$  and 
${\lambda}$ is the regularization parameter.

\subsection{Semi-Supervised Learning Approach.}
Semi-Supervised Learning (SSL) involves
use of both labeled and unlabeled data for learning 
the parameters of the classification model. SSL 
approaches lie between unsupervised (no labeled training data) 
and supervised learning (completely 
labeled training data) \cite{41}. SSL works
on the principle that more the training examples leads to 
better generalization. However, the 
performance of SSL is largely dependent on how 
we treat the unlabeled data with the 
labeled data. 

Our SSL approach works the same way as the
STL method with only difference in the increase in 
number of labeled examples from the related classes found 
using the kNN approach.
Within the SSL approach, for each classification task 
we treat training examples from related classes as positive 
examples for the class under consideration.
We implemented the SSL approach along with STL approach as baseline
to compare against the developed 
MTL and TL approaches.

\section{Experimental Evaluations}
\subsection{Dataset}
To evaluate our methods, we used DMOZ and Wikipedia 
datasets from ECML/PKDD 2012 Large Scale Hierarchical 
Text Classification Challenge (LSHTC) (Track 2) \footnote{http://lshtc.iit.demokritos.gr/LSHTC3\_DATASETS}. The
challenge is closed for new 
submission and the labels of the test set
are not publicly available.  We used the original
training set for training, validation and 
testing by splitting it into 3:1:1 ratios, respectively and reporting the average of five runs. To
assess the performance of our 
method with respect to the class size, in terms of 
the number of training examples, we categorized the classes 
into Low Distribution (LD), with 25 examples per class
and High Distribution (HD), with 250 examples per class.
This resulted in DMOZ dataset having 75 classes within LD and 
53 classes within HD. For the Wikipedia dataset we had 
84 classes within LD and 62 classes within HD. More details about the dataset can be found in the Naik et. al. thesis \cite{naik2013using}.

\subsection{Implementation}

For learning the weight vectors across all the models, we implemented
gradient descent algorithm. Implementation was done in MATLAB and all runs
were performed on a server workstation with a dual-core 
Intel Xeon CPU 2.40GHz processor  and 4GB RAM.

\subsection{Metrics\label{sec:metrics}}
We 
used three 
 metrics for evaluating 
 the classification performance
 that take into account True Positives ($TP$), False
 Positives ($FP$), True Negatives ($TN$) and False 
 Negatives ($FN$) for each of the classes.

\subsubsection{Micro-Averaged $F_1$}
Micro-Averaged $F_1$ ($\mu$A$F_1$) is a conventional metric for evaluating classifiers in category assignment\cite{39}\cite{40}. To compute this metric we sum up the category specific True Positives $(TP_c)$, False Positives $(FP_c)$, True Negatives $(TN_c)$ and False Negatives $(FN_c)$ across all the categories, $c\in C \equiv \{c_1,c_2,\ldots,c_{N_c}\}$ and compute the averaged $F_1$ score. It is defined as follows,

\begin{equation}Global Precision \;P = \frac{\sum_{c=1}^{N_c}TP_c}{\sum_{c=1}^{N_c}(TP_c + FP_c)}\end{equation}
\begin{equation}Global Recall \;R = \frac{\sum_{c=1}^{N_c}TP_c}{\sum_{c=1}^{N_c}(TP_c + FN_c)}\end{equation}
\begin{equation}\mu A F_1 = \frac{2PR}{P + R}\end{equation}
where, $N_c$ is the number of categories/classes.\\

\subsubsection{Macro-Averaged Precision, Recall and $F_1$}
The Macro-Averaged Precision (MAP), Recall (MAR) and $F_1$ (MA$F_1$) are computed by calculating the respective Precision, Recall and $F_1$ scores for each individual category and then averaging them across all the categories\cite{37}. In computing these  metrics all the categories are given equal weight so that the score is not skewed in favor of the larger categories,

\begin{equation}Category-specific \;Precision \;P_c = \frac{TP_c}{TP_c + FP_c}\end{equation}
\begin{equation}Category-specific \;Recall \;R_c = \frac{TP_c}{TP_c + FN_c}\end{equation}
\begin{equation}MAP = \frac{1}{N_c}\sum_{c=1}^{N_c}\frac{TP_c}{TP_c + FP_c}\end{equation}
\begin{equation}MAR = \frac{1}{N_c}\sum_{c=1}^{N_c}\frac{TP_c}{TP_c + FN_c}\end{equation}
\begin{equation}M A F_1 = \frac{1}{N_c}\sum_{c=1}^{N_c}\frac{2P_cR_c}{P_c + R_c}\end{equation}\\

\subsubsection{Averaged Matthews Correlation Coefficient score}
Matthews Correlation Coefficient (MCC) score \cite{38} is a balanced measure for binary classification which quantifies the correlation between the actual and predicted values. It returns a value between -1 and +1, where +1 indicates a perfect prediction, a score of 0 signifies no correlation and -1 indicate a perfect negative correlation between the actual and predicted values. The category specific MCC and averaged MCC are defined as,

\begin{multline} MCC_c = \\ \frac{(TP_c*TN_c) - (FP_c*FN_c)}{\sqrt{(TP_c + FP_c)(TP_c +  FN_c)(TN_c + FP_c)(TN_c + FN_c)}}\end{multline}
\begin{equation} Avg. \; MCC \; (AMCC) = \frac{1}{N_c}\sum_{c=1}^{N_c} MCC_c \end{equation}

\section{Results}

\begin{figure}
\centering
\includegraphics[scale=0.4]{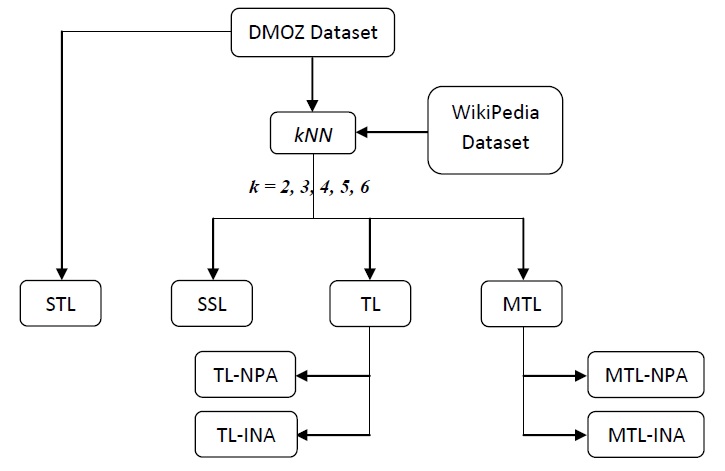}
\caption{Summary of Evaluated Models. Wikipedia/DMOZ are inter-changeable in the protocol.} 
\label{fig:schema}
\end{figure}

We have implemented different models described in 
Section III using  DMOZ and Wikipedia as the two source datasets. 
Figure \ref{fig:schema} outlines the different models that were evaluated.  We varied 
$k$ (number of nearest neighbor) from 2 to 6. 

\begin{table*}
\begin{centering}
\begin{tabular}{|cc||cc|cc|cc|cc|cc|}
\hline \hline
Model & & \multicolumn{2}{c|}{$\mu$A$F_1$} & \multicolumn{2}{c|}{MAP} & \multicolumn{2}{c|}{MAR} & \multicolumn{2}{c|}{MA$F_1$} &  \multicolumn{2}{c|}{AMCC}\\ \hline \hline 
STL && 0.5758 & (0.0121) & 0.7914 & (0.0267) & 0.6167 & (0.0115) & 0.6486 & (0.0060) & 0.6732 & (0.0074)\\ \hline \hline
\multirow{5}{*}{SSL} & ($k = 2$) & 0.6178 & (0.0160) & 0.8064 & (0.0413) & 0.6649 & (0.0125) & 0.6789  & (0.0231)  & 0.7048  & (0.0263) \\
& ($k = 3$) & 0.6316 & (0.0216) & 0.8059 & (0.0218) & 0.6782 & (0.0230) & 0.6808 & (0.0195) & 0.7092 & (0.0185)\\
& ($k = 4$) & 0.6581 & (0.0392) & 0.8090 & (0.0292) & 0.6888 & (0.0329) & 0.6974 & (0.0284) & 0.7208 & (0.0492)\\
& ($k = 5$) & 0.6719 & (0.0540) & \bf0.8091 & \bf(0.0406) & 0.7093 & (0.0314) & 0.7045 & (0.0152) & 0.7359 & (0.0064) \\ 
&  ($k = 6$) & 0.6624 & (0.0356) & 0.8011 & (0.0216) & 0.7089 & (0.0387) & 0.6919 & (0.0284) & 0.7175 & (0.0182)\\ \hline \hline
\multirow{5}{*}{TL-NPA} & ($k = 2$) & 0.5739 & (0.0044) & 0.7987 & (0.0278) & 0.6247 & (0.0091) & 0.6481 & (0.0062)  & 0.6732 & (0.0074) \\
& ($k = 3$) & 0.5728 & (0.0103) & 0.7893 & (0.0128) & 0.6250 & (0.0126) & 0.6480 & (0.0081) & 0.6737 & (0.0171)\\
& ($k = 4$) & 0.5732 & (0.0124) & 0.7981 & (0.0221) & 0.6232 & (0.0753) & 0.6483 & (0.0375) & 0.6738 & (0.0128)\\
& ($k = 5$) & 0.5755 & (0.0030) & 0.8027 & (0.0314) & 0.6262 & (0.0077) & 0.6504 & (0.0083) & 0.6757 & (0.0097)\\ 
& ($k = 6$) & 0.5738 & (0.0462) & 0.7413 & (0.0593) & 0.6096 & (0.0522) & 0.6248 & (0.0563) & 0.6631 & (0.0387)\\ \hline \hline
\multirow{5}{*}{TL-INA} & ($k = 2$) & 0.5736 & (0.0038) & 0.7967 & (0.0262) & 0.6246 & (0.0184) & 0.6478 & (0.0040)  &  0.6728 & (0.0054) \\ 
& ($k = 3$) & 0.5810 & (0.0731) & 0.7918 & (0.0137) & 0.6182 & (0.0126) & 0.6488 & (0.0031) & 0.6794 & (0.0138)\\
& ($k = 4$) & 0.5771 & (0.0192) & 0.7939 & (0.0191) & 0.6173 & (0.0188) & 0.6394 & (0.0113) & 0.6748 & (0.0312)\\
& ($k = 5$) & 0.5712 & (0.0034) & 0.8024 & (0.0226) & 0.6212 & (0.0091) & 0.6467 & (0.0034) & 0.6724 & (0.0047)\\ 
& ($k = 6$) & 0.5700 & (0.0144) & 0.7853 & (0.0268) & 0.6132 & (0.0164) & 0.6298 & (0.0372) & 0.6489 & (0.0189)\\ \hline \hline
\multirow{5}{*}{MTL-NPA} & ($k = 2$) & \bf0.7442 & \bf(0.0201) & 0.7819 & (0.0356) & \bf0.7840 & \bf(0.0169) & \bf0.7373  & \bf(0.0349) & \bf0.7527 & \bf(0.0335)\\ 
& ($k = 3$) & 0.7438 & (0.0192) & 0.7901 & (0.0461) & 0.7782 & (0.0329) & 0.7350 & (0.0247) & 0.7515 & (0.0282)\\ 
& ($k = 4$) & 0.7403 & (0.0431) & 0.7884 & (0.0453) & 0.7891 & (0.0212) & 0.7346 & (0.0221) & 0.7501 & (0.0101)\\  
& ($k = 5$) & 0.7394 & (0.0219) & 0.7720 & (0.0421) & 0.7814 & (0.0140) & 0.7293 & (0.0318) & 0.7488 & (0.0298)\\ 
& ($k = 6$) & 0.7120 & (0.0128) & 0.7104 & (0.0144) & 0.7581 & (0.0213) & 0.6866 & (0.0422) & 0.7061 & (0.0431)\\ \hline \hline
\multirow{5}{*}{MTL-INA} & ($k = 2$) & 0.7208 & (0.0180) & 0.7583 & (0.0503) & 0.7664 & (0.0211) & 0.7052 & (0.0520) & 0.7326 & (0.0367)  \\
& ($k = 3$) & 0.7294 & (0.0213) & 0.7592 & (0.0101) & 0.7616 & (0.0473) & 0.7070 & (0.0211) & 0.7313 & (0.0312)\\
& ($k = 4$) & 0.7000 & (0.0431) & 0.7281 & (0.0213) & 0.7502 & (0.0131) & 0.6899 & (0.0432) & 0.7024 & (0.0293)\\
& ($k = 5$) & 0.7079 & (0.0136) & 0.7352 & (0.0306) & 0.7508 & (0.0085) & 0.6949 & (0.0255) & 0.7147 & (0.0243)\\ 
& ($k = 6$) & 0.6992 & (0.0721) & 0.7271 & (0.0721) & 0.7321 & (0.0632) & 0.6797 & (0.0413) & 0.7024 & (0.0339)\\ \hline \hline
\end{tabular}
\par\end{centering}
    \begin{tablenotes}
      \tiny
      \item \center Table shows mean across five runs and (standard deviation) in bracket, standard error for best model: MTL-NPA ($k$ = 2) = 0.0054
    \end{tablenotes}
\caption{Classification Performance for LD Sample shown for DMOZ dataset. \label{tab:low-distribution-DMOZ}}
\end{table*}

\begin{table*}
\begin{centering}
\begin{tabular}{|cc||cc|cc|cc|cc|cc|}
\hline \hline
Model & & \multicolumn{2}{c|}{$\mu$A$F_1$} & \multicolumn{2}{c|}{MAP} & \multicolumn{2}{c|}{MAR} & \multicolumn{2}{c|}{MA$F_1$} &  \multicolumn{2}{c|}{AMCC}\\ \hline \hline 
STL && 0.5236 & (0.0989) & 0.6415 & (0.0741) & 0.5213 & (0.0998) & 0.5318 & (0.0625) & 0.5837 & (0.0782)\\ \hline \hline
\multirow{5}{*}{SSL} & ($k = 2$) & 0.5182 & (0.0321) & 0.6392 & (0.0932) & 0.5348 & (0.0631) & 0.5264 & (0.0674) & 0.5748 & (0.0183)\\
& ($k = 3$) & 0.5234 & (0.0723) & 0.6334 & (0.0673) & 0.5354 & (0.0300) & 0.5670 & (0.0641) & 0.5739 & (0.0629)\\
& ($k = 4$) & 0.5329 & (0.0631) & 0.6312 & (0.0681) & 0.5360 & (0.0810) & 0.5698 & (0.0524) & 0.5802 & (0.0285)\\
& ($k = 5$) & 0.5102 & (0.0642) & 0.6124 & (0.0773) & 0.5279 & (0.0273) & 0.5490 & (0.0831) & 0.5772 & (0.0641)\\ 
& ($k = 6$) & 0.5043 & (0.0731) & 0.6042 & (0.0204) & 0.5186 & (0.0942) & 0.5468 & (0.0632) & 0.5547 & (0.0228)\\ \hline \hline
\multirow{5}{*}{TL-NPA} & ($k = 2$) & 0.5418 & (0.0182) & 0.6682 & (0.0136) & 0.5633 & (0.0362) & 0.5823 & (0.0317)  & 0.6930 & (0.0831) \\
& ($k = 3$) & 0.5512 & (0.0521) & 0.6620 & (0.0317) & 0.5784 & (0.0674) & 0.5982 & (0.0742) & 0.6894 & (0.0674)\\
& ($k = 4$) & 0.5332 & (0.0153) & 0.6581 & (0.0239) & 0.5616 & (0.0083) & 0.5813 & (0.0873) & 0.6740 & (0.0543)\\
& ($k = 5$) & 0.5295 & (0.0743) & 0.6327 & (0.0854) & 0.5610 & (0.0674) & 0.5704 & (0.0029) & 0.6649 & (0.0895)\\ 
& ($k = 6$) & 0.5238 & (0.0235) & 0.6136 & (0.0487) & 0.5427 & (0.0500) & 0.5684 & (0.0748) & 0.6386 & (0.0856)\\ \hline \hline
\multirow{5}{*}{TL-INA} & ($k = 2$) & 0.5368 & (0.0573) & 0.6734 & (0.0198) & 0.5464 & (0.0563) & 0.5982 & (0.0130)  &  0.6946 & (0.0846) \\ 
& ($k = 3$) & 0.5408 & (0.0464) & 0.6648 & (0.0895) & 0.5696 & (0.0187) & 0.5928 & (0.0481) & 0.6994 & (0.0101)\\
& ($k = 4$) & 0.5319 & (0.0042) & 0.6598 & (0.0452) & 0.5573 & (0.0526) & 0.5810 & (0.0736) & 0.6848 & (0.0654)\\
& ($k = 5$) & 0.5278 & (0.0674) & 0.6394 & (0.0895) & 0.5210 & (0.0183) & 0.5624 & (0.0901) & 0.6740 & (0.0538)\\ 
& ($k = 6$) & 0.5101 & (0.0587) & 0.6153 & (0.0519) & 0.5052 & (0.0456) & 0.5248 & (0.0831) & 0.6382 & (0.0873)\\ \hline \hline
\multirow{5}{*}{MTL-NPA} & ($k = 2$) & 0.6389 & (0.0648) & 0.6635 & (0.0782) & 0.6615 & (0.0637) & 0.6626 & (0.0682) & 0.6582 & (0.0738)\\ 
& ($k = 3$) & \bf0.6390 & \bf(0.0723) & \bf0.6832 & \bf(0.0672) & \bf0.6650 & \bf(0.0421) & \bf0.6724 & \bf(0.0432) & \bf0.6628 & \bf(0.0764)\\ 
& ($k = 4$) & 0.6283 & (0.0748) & 0.6624 & (0.0613) & 0.6593 & (0.0382) & 0.6602 & (0.0936) & 0.6585 & (0.0631)\\  
& ($k = 5$) & 0.6128 & (0.0784) & 0.6626 & (0.0632) & 0.6429 & (0.0823) & 0.6593 & (0.0529) & 0.6497 & (0.0874)\\ 
& ($k = 6$) & 0.6003 & (0.0524) & 0.6498 & (0.0874) & 0.6193 & (0.0623) & 0.6282 & (0.0817) & 0.6046 & (0.0734)\\ \hline \hline
\multirow{5}{*}{MTL-INA} & ($k = 2$) & 0.6120 & (0.0629) & 0.6448 & (0.0325) & 0.6428 & (0.0618) & 0.6432 & (0.0663) & 0.6420 & (0.0728)\\
& ($k = 3$) & 0.6194 & (0.0437) & 0.6328 & (0.0224) & 0.6480 & (0.0642) & 0.6406 & (0.0910) & 0.6394 & (0.0429)\\
& ($k = 4$) & 0.6036 & (0.0421) & 0.6262 & (0.0639) & 0.6338 & (0.0632) & 0.6324 & (0.0138) & 0.6310 & (0.0309)\\
& ($k = 5$) & 0.6040 & (0.0101) & 0.6160 & (0.0819) & 0.6282 & (0.0192) & 0.6202 & (0.0328) & 0.6290 & (0.0456)\\ 
& ($k = 6$) & 0.5842 & (0.0457) & 0.5820 & (0.0282) & 0.5926 & (0.0478) & 0.5846 & (0.0885) & 0.6082 & (0.0402)\\ \hline \hline
\end{tabular}
\par\end{centering}
    \begin{tablenotes}
      \tiny
      \item \center Table shows mean across five runs and (standard deviation) in bracket, standard error for best model: MTL-NPA ($k$ = 3) = 0.0087
    \end{tablenotes}
\caption{Classification Performance for LD sample shown for Wikipedia dataset. \label{tab:low-distribution-Wikipedia}}
\end{table*}


\subsection{Accuracy Comparison}
\subsubsection{Low distribution sample}
Tables \ref{tab:low-distribution-DMOZ} and \ref{tab:low-distribution-Wikipedia} show the average performance across five runs for DMOZ and Wikipedia Low Distribution (LD) classes. The following observations can be made from the results.
\begin{itemize}
\item {\bf {STL v/s SSL v/s TL v/s MTL}}:
For both the datasets we see that the 
MTL methods outperforms all the 
other methods across all the metrics, exception being in case of LD DMOZ dataset MAP metric. Reason for such exception is high relatedness between the main task and its corresponding neighboring task(s). We also 
note that among the MTL approaches the 
Neighborhood Pooling Approach (MTL-NPA)   outperformed 
the Individual Neighborhood Approach (MTL-INA) (statistically significant). Semi-Supervised Learning (SSL) method
marginally outperformed both Single Task Learning (STL)
as well as Transfer Learning (TL) methods. TL did not seem 
to have any benefit over STL. 
\item {\bf {$k$ = 2 v/s $k$ = 3 v/s $k$ = 4 v/s $k$ = 5 v/s $k$ = 6}}:
In general, lower value of $k$ gave better models compared to higher values of $k$. We conjecture that as the value of $k$ increases, similarity between the main task and the surrogate tasks decreases, which in turn affects the performance negatively.
\end{itemize}

\begin{table*}
\begin{centering}
\begin{tabular}{|cc||cc|cc|cc|cc|cc|}
\hline \hline
Model & & \multicolumn{2}{c|}{$\mu$A$F_1$} & \multicolumn{2}{c|}{MAP} & \multicolumn{2}{c|}{MAR} & \multicolumn{2}{c|}{MA$F_1$} &  \multicolumn{2}{c|}{AMCC}\\ \hline \hline 
STL && 0.7592 & (0.0120) & 0.7947 & (0.0068) & 0.7618 & (0.0056) & 0.7567 & (0.0085) & 0.7626 & (0.0075)\\ \hline \hline
\multirow{5}{*}{SSL} & ($k = 2$) & 0.7535 & (0.0040) & 0.7938 & (0.0065) & 0.7601 & (0.0011) & 0.7547 & (0.0052) & 0.7609 & (0.0048)\\ 
& ($k = 3$) & 0.7542 & (0.0127) & 0.7998 & (0.0032) & 0.7624 & (0.0102) & 0.7657 & (0.0020) & 0.7646 & (0.0091)\\
& ($k = 4$) & 0.7512 & (0.0031) & 0.7972 & (0.0084) & 0.7618 & (0.0090) & 0.7592 & (0.0029) & 0.7628 & (0.0075)\\
& ($k = 5$) & 0.7545 & (0.0027) & 0.7948 & (0.0060) & 0.7610 & (0.0018) & 0.7559 & (0.0039) & 0.7619 & (0.0034)\\ 
& ($k = 6$) & 0.7491 & (0.0042) & 0.7778 & (0.0143) & 0.7482 & (0.0042) & 0.7437 & (0.0092) & 0.7542 & (0.0086)\\ \hline \hline
\multirow{5}{*}{TL-NPA} & ($k = 2$) & 0.7536 & (0.0042) & 0.7936 & (0.0054) & 0.7588 & (0.0015) & 0.7546 & (0.0048) & 0.7610 & (0.0043)\\ 
& ($k = 3$) & 0.7538 & (0.0053) & 0.7941 & (0.0027) & 0.7590 & (0.0061) & 0.7551 & (0.0024) & 0.7595 & (0.0063)\\
& ($k = 4$) & 0.7532 & (0.0028) & 0.7940 & (0.0037) & 0.7585 & (0.0072) & 0.7544 & (0.0053) & 0.7623 & (0.0022)\\
& ($k = 5$) & 0.7533 & (0.0038) & 0.7937 & (0.0063) & 0.7584 & (0.0024) & 0.7542 & (0.0052) & 0.7605 & (0.0046)\\ 
& ($k = 6$) & 0.7406 & (0.0064) & 0.7888 & (0.0022) & 0.7414 & (0.0072) & 0.7389 & (0.0041) & 0.7594 & (0.0027)\\ \hline \hline
\multirow{5}{*}{TL-INA} & ($k = 2$) & 0.7529 & (0.0040) & 0.7958 & (0.0054) & 0.7591 & (0.0011) & 0.7540 & (0.0048) & 0.7603 & (0.0043)\\ 
& ($k = 3$) & 0.7631 & (0.0074) & 0.7964 & (0.0010) & 0.7612 & (0.0027) & 0.7548 & (0.0029) & 0.7696 & (0.0072)\\
& ($k = 4$) & 0.7530 & (0.0091) & 0.7968 & (0.0020) & 0.7606 & (0.0013) & 0.7542 & (0.0017) & 0.7596 & (0.0062)\\
& ($k = 5$) & 0.7527 & (0.0038) & 0.7957 & (0.0046) & 0.7590 & (0.0010) & 0.7538 & (0.0036) & 0.7602 & (0.0042)\\ 
& ($k = 6$) & 0.7432 & (0.0015) & 0.7849 & (0.0034) & 0.7442 & (0.0051) & 0.7414 & (0.0053) & 0.7591 & (0.0068)\\ \hline \hline
\multirow{5}{*}{MTL-NPA} & ($k = 2$) & 0.7572 & (0.0080) & 0.7961 & (0.0063) & 0.7637 & (0.0058) & 0.7587 & (0.0087) & 0.7644 & (0.0076)\\ 
& ($k = 3$) & 0.7598 & (0.0072) & 0.7978 & (0.0089) & 0.7644 & (0.0062) & 0.7649 & (0.0074) & 0.7701 & (0.0086)\\
& ($k = 4$) & 0.7571 & (0.0035) & 0.7964 & (0.0076) & 0.7632 & (0.0093) & 0.7584 & (0.0102) & 0.7640 & (0.0054)\\
& ($k = 5$) & 0.7569 & (0.0043) & 0.7969 & (0.0058) & 0.7627 & (0.0012) & 0.7581 & (0.0053) & 0.7639 & (0.0048)\\ 
& ($k = 6$) & 0.7482 & (0.0087) & 0.7712 & (0.0081) & 0.7512 & (0.0014) & 0.7392 & (0.0076) & 0.7436 & (0.0030)\\ \hline \hline
\multirow{5}{*}{MTL-INA} & ($k = 2$) & 0.7579 & (0.0103) & 0.7978 & (0.0076) & 0.7644 & (0.0079) & 0.7599 & (0.0111) & 0.7657 & (0.0099)\\ 
& ($k = 3$) & \bf0.7680 & \bf(0.0097) & \bf0.8020 & \bf(0.0047) & \bf0.7790 & \bf(0.0089) & \bf0.7728 & \bf(0.0092) & \bf0.7728 & \bf(0.0121)\\
& ($k = 4$) & 0.7662 & (0.0087) & 0.8012 & (0.0026) & 0.7742 & (0.0085) & 0.7696 & (0.0129) & 0.7719 & (0.0105)\\
& ($k = 5$) & 0.7657 & (0.0067) & 0.8017 & (0.0073) & 0.7717 & (0.0037) & 0.7678 & (0.0081) & 0.7726 & (0.0072)\\ 
& ($k = 6$) & 0.7526 & (0.0089) & 0.7818 & (0.0129) & 0.7664 & (0.0051) & 0.7529 & (0.0066) & 0.7648 & (0.0059)\\ \hline \hline
\end{tabular}
\par\end{centering}
    \begin{tablenotes}
      \tiny
      \item \center Table shows mean across five runs and (standard deviation) in bracket, standard error for best model: MTL-INA ($k$ = 3) = 0.0038
    \end{tablenotes}
\caption{Classification Performance for HD sample shown for  DMOZ dataset. \label{tab:high-distribution-DMOZ}}
\end{table*}

\begin{table*}
\begin{centering}
\begin{tabular}{|cc||cc|cc|cc|cc|cc|}
\hline \hline
Model & & \multicolumn{2}{c|}{$\mu$A$F_1$} & \multicolumn{2}{c|}{MAP} & \multicolumn{2}{c|}{MAR} & \multicolumn{2}{c|}{MA$F_1$} &  \multicolumn{2}{c|}{AMCC}\\ \hline \hline 
STL && 0.6648 & (0.0628) & 0.6841 & (0.0630) & 0.6429 & (0.0089) & 0.6748 & (0.0172) & 0.6881 & (0.0120)\\ \hline \hline
\multirow{5}{*}{SSL} & ($k = 2$) & 0.6584 & (0.0067) & 0.6848 & (0.0178) & 0.6492 & (0.0238) & 0.6780 & (0.0324) & 0.6782 & (0.0182)\\ 
& ($k = 3$) & 0.6528 & (0.0262) & 0.6804 & (0.0572) & 0.6498 & (0.0546) & 0.6680 & (0.0821) & 0.6778 & (0.0239)\\
& ($k = 4$) & 0.6530 & (0.0287) & 0.6768 & (0.0231) & 0.6400 & (0.0262) & 0.6612 & (0.0387) & 0.6704 & (0.0263)\\
& ($k = 5$) & 0.6428 & (0.0624) & 0.6706 & (0.0189) & 0.6364 & (0.0423) & 0.6596 & (0.0346) & 0.6686 & (0.0456)\\ 
& ($k = 6$) & 0.6320 & (0.0822) & 0.6700 & (0.0037) & 0.6342 & (0.0892) & 0.6502 & (0.0521) & 0.6648 & (0.0190)\\ \hline \hline
\multirow{5}{*}{TL-NPA} & ($k = 2$) & 0.6628 & (0.0636) & 0.6820 & (0.0976) & 0.6528 & (0.0174) & 0.6792 & (0.0733) & 0.6840 & (0.0785)\\ 
& ($k = 3$) & 0.6614 & (0.0463) & 0.6838 & (0.0842) & 0.6510 & (0.0597) & 0.6797 & (0.0471) & 0.6888 & (0.0823)\\
& ($k = 4$) & 0.6528 & (0.0367) & 0.6735 & (0.0963) & 0.6482 & (0.0871) & 0.6626 & (0.0913) & 0.6710 & (0.0731)\\
& ($k = 5$) & 0.6500 & (0.0689) & 0.6629 & (0.0729) & 0.6285 & (0.0463) & 0.6389 & (0.0582) & 0.6618 & (0.0838)\\ 
& ($k = 6$) & 0.6450 & (0.0893) & 0.6482 & (0.0572) & 0.6021 & (0.0578) & 0.6124 & (0.0527) & 0.6484 & (0.0657)\\ \hline \hline
\multirow{5}{*}{TL-INA} & ($k = 2$) & 0.6531 & (0.0462) & 0.6623 & (0.0572) & 0.6482 & (0.0863) & 0.6504 & (0.0427) & 0.6731 & (0.0865)\\ 
& ($k = 3$) & 0.6512 & (0.0845) & 0.6547 & (0.0864) & 0.6273 & (0.0974) & 0.6397 & (0.0645) & 0.6682 & (0.0472)\\
& ($k = 4$) & 0.6510 & (0.0467) & 0.6524 & (0.0246) & 0.6244 & (0.0755) & 0.6326 & (0.0624) & 0.6539 & (0.0573)\\
& ($k = 5$) & 0.6427 & (0.0533) & 0.6510 & (0.0217) & 0.6218 & (0.0381) & 0.6304 & (0.0256) & 0.6512 & (0.0972)\\ 
& ($k = 6$) & 0.6308 & (0.0572) & 0.6404 & (0.0384) & 0.6036 & (0.0330) & 0.6198 & (0.0472) & 0.6380 & (0.0753)\\ \hline \hline
\multirow{5}{*}{MTL-NPA} & ($k = 2$) & 0.6620 & (0.0672) & 0.6824 & (0.0317) & 0.6428 & (0.0871) & 0.6704 & (0.0174) & 0.6868 & (0.0623)\\ 
& ($k = 3$) & 0.6702 & (0.0053) & 0.6826 & (0.0183) & 0.6440 & (0.0542) & 0.6748 & (0.0831) & 0.6880 & (0.0542)\\
& ($k = 4$) & 0.6634 & (0.0184) & 0.6782 & (0.0172) & 0.6210 & (0.0281) & 0.6529 & (0.0600) & 0.6693 & (0.0622)\\
& ($k = 5$) & 0.6608 & (0.0731) & 0.6616 & (0.0722) & 0.6201 & (0.0193) & 0.6500 & (0.0783) & 0.6524 & (0.0734)\\ 
& ($k = 6$) & 0.6529 & (0.0620) & 0.6583 & (0.0318) & 0.6183 & (0.0731) & 0.6472 & (0.0561) & 0.6510 & (0.0582)\\ \hline \hline
\multirow{5}{*}{MTL-INA} & ($k = 2$) & \bf0.6720 & \bf(0.0134) & \bf0.6898 & \bf(0.0531) & 0.6548 & (0.0146) & \bf0.6784 & \bf(0.0142) & \bf0.6898 & \bf(0.0712)\\ 
& ($k = 3$) & 0.6717 & (0.0108) & 0.6864 & (0.0142) & \bf0.6550 & \bf(0.0398) & 0.6772 & (0.0152) & 0.6720 & (0.0256)\\
& ($k = 4$) & 0.6683 & (0.0040) & 0.6747 & (0.0051) & 0.6484 & (0.0193) & 0.6696 & (0.0641) & 0.6704 & (0.0138)\\
& ($k = 5$) & 0.6601 & (0.0839) & 0.6630 & (0.0931) & 0.6418 & (0.0322) & 0.6642 & (0.0412) & 0.6652 & (0.0313)\\ 
& ($k = 6$) & 0.6539 & (0.0713) & 0.6565 & (0.0172) & 0.6402 & (0.0742) & 0.6598 & (0.0193) & 0.6584 & (0.0105)\\ \hline \hline
\end{tabular}
\par\end{centering}
    \begin{tablenotes}
      \tiny
      \item \center Table shows mean across five runs and (standard deviation) in bracket, standard error for best model: MTL-INA ($k$ = 2) = 0.0068
    \end{tablenotes}
\caption{Classification Performance for HD sample shown for Wikipedia dataset. \label{tab:high-distribution-Wikipedia}}
\end{table*}
\subsubsection{High distribution sample}
Table \ref{tab:high-distribution-DMOZ} and \ref{tab:high-distribution-Wikipedia} show the average performance across five runs for DMOZ and Wikipedia High Distribution (HD) classes. We make the following observations based on the results.

\begin{itemize}
\item {\bf {STL v/s SSL v/s TL v/s MTL}}: 
In this case we see that the MTL methods perform only slightly better than other models, the differences are not statistically significant. This supports our intuition that with sufficient number of examples for learning the SSL, TL and MTL methods do not provide any distinct advantage and the simple STL model is 
competitive. 
\item {\bf {$k$ = 2 v/s $k$ = 3 v/s $k$ = 4 v/s $k$ = 5 v/s $k$ = 6}}:
As with the case of Low Distribution classes, we noticed a slight degradation of performance as the number of neighbors is increased. 
\end{itemize}


\subsection{Run time Comparison}
Table \ref{tab:runtime-DMOZ} shows the average training 
time (in sec.) per class required to
learn the models for the different
LD and HD categories. The STL approach
has the lowest training times because there is 
no overhead  of incorporating additional 
constraints is involved. SSL models takes more time than the 
corresponding STL models because of the increased
number of training examples. For TL models as 
well, run time increases because it requires learning the
models for the neighbors. Finally, MTL method takes the longest time, since
it requires 
the joint learning of the model parameters
that are updated for each class and related neighbors.

\begin{table}
\begin{small}
\begin{centering}
\begin{tabular}{|cc||cc|cc|}
\hline \hline
& & \multicolumn{2}{c}{DMOZ} &\multicolumn{2}{c|}{Wiki}\\\hline
Model & & LD & HD & LD & HD\\ \hline \hline 
STL && 2.72 & 44.7 & 2.84 & 46.7\\ \hline \hline
\multirow{3}{*}{SSL} & ($k = 2$) & 4.58 & 62.4 & 8.64 &68.3\\
& ($k = 4$) & 5.57 & 62.7 & 9.73 & 70.5\\
& ($k = 6$) & 6.48 & 64.3 & 10.6 & 73.0 \\ \hline \hline
\multirow{3}{*}{TL-NPA} & ($k = 2$) & 4.65 & 48.6 & 10.2 & 70.7 \\
& ($k = 4$) & 6.3 & 52.6 & 13.6  & 74.6 \\
& ($k = 6$) & 7.5 & 56.1 & 15.5 & 78.8  \\ \hline \hline
\multirow{3}{*}{TL-INA} & ($k = 2$) & 4.54 & 48.4 & 12.6 & 72.6 \\
& ($k = 4$) & 6.40 & 50.3 & 14.6 & 76.8\\
& ($k = 6$) & 7.98 & 54.7 & 15.6 & 80.7\\ \hline \hline
\multirow{3}{*}{MTL-NPA} & ($k = 2$) & 5.51 & 49.5 & 12.6  & 69.6 \\
& ($k = 4$) & 7.53 & 54.2 & 14.5  & 70.3\\
& ($k = 6$) & 8.75 & 58.1 & 15.8 & 72.4\\ \hline \hline
\multirow{3}{*}{MTL-INA} & ($k = 2$) & 9.84 & 56.8 & 17.8 & 78.7\\
& ($k = 4$) & 15.6  & 78.7 & 19.3 & 84.7\\
& ($k = 6$) & 18.8 & 82.8  & 22.3 & 92.4 \\ \hline \hline 
\end{tabular}
\par\end{centering}
\caption{Run time (in sec.) comparison (Reported per class across five runs)\label{tab:runtime-DMOZ}}
\end{small}
\end{table}
%

\section{Conclusion and Future Work}
In this paper we developed Multi-task Learning models for text document classification. Performance of the MTL methods was compared with Single Task Learning, Semi-supervised Learning and Transfer Learning approaches. We compared the methods in terms of accuracy and run-times. MTL methods outperformed the other methods, especially for the Low Distribution classes, where the number of positive training examples was small. For the High Distribution classes with sufficient number of positive training examples, the performance improvement was not noticeable. 

Datasets organize information as hierarchies. We plan to 
extract the parent-child relationships existing within the DMOZ and Wikipedia hierarchies 
to improve the classification performance.
We also plan to use the accelerated/proximal gradient descent approach to improve the
learning rates. Finally, we also seek to improve run-time performance
by implementing our approaches using data parallelism, seen in GPUs.

\section{Acknowledgement}
This project was funded by NSF career award 1252318.

\bibliographystyle{unsrt}
\bibliography{MTLTextClassification} 

\end{document}